\documentclass[conference]{IEEEtran}
\IEEEoverridecommandlockouts
\usepackage{cite}
\usepackage{amsmath,amssymb,amsfonts}
\usepackage{algorithmic}
\usepackage{graphicx}
\usepackage{textcomp}
\usepackage{xcolor}
\usepackage{amsthm}
 
\theoremstyle{definition}
\newtheorem{definition}{Definition}
\theoremstyle{definition} 
\theoremstyle{remark}

\usepackage{booktabs} 
\usepackage{multicol}  
\usepackage{multirow}  
\usepackage{algorithm}
\usepackage{amsfonts} 
\usepackage{amsmath}
\usepackage{amssymb}

\def\BibTeX{{\rm B\kern-.05em{\sc i\kern-.025em b}\kern-.08em
    T\kern-.1667em\lower.7ex\hbox{E}\kern-.125emX}}
\begin{document}

\author{
\IEEEauthorblockN{Yiwei Sun$^{\dag,\S}$, Ngot Bui$^{\ddag}$, Tsung-Yu Hsieh$^{\dag,\S}$ and Vasant Honavar$^{\dag,\ast}$}
\IEEEauthorblockA{$^{\dag}$Artificial Intelligence Research Laboratory,  The Pennsylvania State University, USA\\
$^{\ast}$College of Information Sciences and Technology, The Pennsylvania State University, USA\\
$^{\S}$Department of Computer Science and Engineering, The Pennsylvania State University, USA\\
$^{\ddag}$Google LLC, USA\\
Email: yus162@psu.edu, bpngot@google.com, tuh45@psu.edu, vhonavar@ist.psu.edu\\
}}

\title{Multi-View Network Embedding Via Graph Factorization Clustering and Co-Regularized Multi-View Agreement}

\maketitle

\begin{abstract}
Real-world social networks and digital platforms are comprised of individuals (nodes) that are linked to other individuals or entities 
through multiple types of relationships (links).  Sub-networks of such a network based on each type of link correspond to distinct views of the underlying network. In real-world applications, each node is typically linked to only a small subset of other nodes. Hence, practical approaches to problems such as node labeling have to cope with the resulting sparse networks.  While low-dimensional network embeddings offer a promising approach to this problem, most of the current network embedding methods focus primarily on single view networks.  We introduce a novel multi-view network embedding (MVNE) algorithm for constructing low-dimensional node embeddings from multi-view networks. MVNE adapts and extends an approach to single view network embedding (SVNE) using graph factorization clustering (GFC) to the multi-view setting using an objective function that maximizes the agreement between views based on both the local and global structure of the underlying multi-view graph. Our experiments with several benchmark real-world single view networks show that GFC-based SVNE yields network embeddings that are  competitive with or superior to those produced by the state-of-the-art single view network embedding methods when the embeddings are used for labeling unlabeled nodes in the networks. Our experiments with several multi-view networks show that MVNE substantially outperforms the single view methods on integrated view and the state-of-the-art multi-view methods. We further show that even when the goal is to predict labels of nodes within a single target view, MVNE outperforms its single-view counterpart suggesting that the MVNE is able to extract the information that is useful for labeling nodes in the target view from the all of the views. 
\end{abstract}
\section{Introduction}
Social networks e.g., Facebook, social media e.g., Flickr, and e-commerce platforms, e.g., Amazon, can be seen as very large heterogeneous networks where the nodes correspond to diverse types of entities, e.g., articles, images, videos, music, etc. In such networks, an individual can link to multiple other individuals via different types of social or other relationships e.g., friendship, co-authorship, etc\cite{dickison2016multilayer,tang2015line,berlingerio2013multidimensional}. Examples include Google+ which allows members to specify different 'circles' that correspond to different types of social relationships;  DBLP which contains multiple types of relationships that link authors to articles, publication venues, institutions, etc.  Such networks are naturally represented as \emph{multi-view} networks wherein the nodes denote individuals  and links denote relationships such that each \emph{network view} corresponds to a single type of relationship, e.g., friendship, family membership, etc\cite{kivela2014multilayer,bazzi2016community,scott2011social,bui2016labeling}. Such networks present several problems of interest, e.g., recommending products, activities or membership in specific interest groups to individuals based on the attributes of individuals, the multiple relationships that link them to entities or other individuals, etc. \cite{elkahky2015multi,benton2016learning}. 

When multiple sources of data are available about entities of interest, multi-view learning  offers a promising approach to integrating complementary information provided by the different data sources (views) to optimize the performance of predictive models \cite{xu2013survey,sun2013survey}.  Examples of such multi-view learning algorithms include: multi-view support vector machines \cite{cao2014tensor, li2010two},  multi-view matrix (tensor) factorization \cite{lu2017multilinear,lu2018learning},  and  multi-view clustering  via  canonical  correlation  analysis \cite{chaudhuri2009multi,dhillon2011multi}. However, most of the existing multi-view learning algorithms are not (i) directly applicable to multi-view networks; and (ii) designed to cope with data sparsity, which is one of the key challenges in modeling real-world multi-view networks: although the number of nodes in real-world networks is often in the millions, typically each node is linked to only a small subset of other nodes.  Low-dimensional network embeddings offer a promising approach to dealing with such sparse networks \cite{cui2017survey}. However, barring a few exceptions \cite{shi2018mvn2vec,ma2017multi,qu2017attention,bui2016labeling}, most of the work on network embedding has focused on methods for single view networks \cite{tang2015line,perozzi2014deepwalk,grover2016node2vec}. 

Against this background, the key contributions of this paper are as follows:
\begin{enumerate} 
\item We introduce a novel multi-view network  embedding (MVNE) algorithm for constructing low-dimensional embeddings of nodes in multi-view networks. MVNE exploits recently discovered connection between network adjacency matrix factorization and network embedding \cite{qiu2018network}. Specifically, we use the graph factorization clustering (GFC) \cite{yu2006soft} algorithm to obtain single view network embedding. MVNE extends the resulting single view network node embedding algorithm (SVNE) to the multi-view setting. Inspired by \cite{lai2017prune}, MVNE integrates both local and global context of nodes in  networks to construct  effective  embeddings of multi-view networks.  Specifically, MVNE uses a novel objective function that maximizes the agreement between views based on both the local and global structure of the underlying multi-view graph.  
\item We present results of experiments  with several benchmark real-world data that demonstrate the effectiveness of MVNE relative to state-of-the-art network embedding methods. Specifically, we show that (i)  SVNE is competitive with or superior to the state-of-the-art single view graph embedding methods when the embeddings are used for labeling unlabeled nodes in single view networks. (ii)  MVNE substantially outperforms the state-of-the-art single view and multi-view embedding methods for aggregating information from multiple views, when the embeddings are used for labeling nodes in multi-view networks. (iii) MVNE is able to augment information from any target  view with relevant information extracted from other views so as to improve node labeling performance on the target view in multi-view networks. 
\end{enumerate}

The rest of the paper is organized as follows. In Section 2, we formally define the problem of multi-view network  embedding. In Section 3, we describe the proposed MVNE framework. In Section 4, we present results of experiments that compare the performance of MVNE with state-of-the-art single view network node embedding methods and their multi-view extensions.  In Section 5, we conclude with a summary, discussion of related work, and some directions for further research. 

\section{Preliminaries}
\begin{definition}(Multi-view Network) A multi-view network is defined by 6-tuple $G = (V,E,T_V,T_E,\phi_V,\phi_E)$ where $V$ is a set of nodes, $E$ is a set of edges, $T_V$ and $T_E$ respectively denote sets of node and relation types, and $\phi_V:V \rightarrow {\mathcal P}(T_V)$ and $\phi_E: E \rightarrow  T_E$ (where ${\mathcal P}(S)$ is the power set of set $S$),  are functions that associate  each node $v \in V$ with a subset of types in $T_V$ and each edge $e \in E$ with their corresponding  type in  $T_E$ respectively.
\end{definition}

Note that a node can have multiple types. For example, in an academic network with nodes types authors (A), professors (R), papers (P), venues (V), organizations (O), topics (T), relation types may denote the coauthor (A-A), publish (A-P), published-in (P-V), has-expertise (R-T), and affiliation (O-A) relationships. An individual in an academic network can be an author, professor, or both.

Note that the node types are selected from the set $V$ of nodes $|T_V|$ (potentially overlapping) subsets $V^{(1)}, V^{(2)} \cdots V^{(|T_V|)}$. Each view of a multi-view network is represented by an adjacency matrix for each type of edge $t \in T_E$. For an edge type that denotes relationships between nodes in $V^{(i)}$, the corresponding adjacency matrix $W^{(t)}$ will be of size $|V^{(i)}|\times |V^{(i)}|$. Thus, a multi-view network $G$ can be represented by a set of single view networks $G^{(1)} \cdots G^{(|T_E|)}$ where $G^{(t)}$ is represented by the adjacency matrix $W^{(t)}$.

\begin{definition} (Node label prediction problem) Suppose we are given a multi-view network $G$ in which only some of the nodes of each node type $t \in T_V$ are assigned a finite subset of labels in $L_t$, where $L_t$ is the set of possible labels for nodes of type $t$. Given such a network $G$, node label prediction entails completing the labeling of $G$, that is,  for each node of type $t$ that does not already have a label $l \in L_t$, specifying whether it should be labeled with $l$ based on the information provided by the nodes and edges of the multi-view network $G$. 
\end{definition}
In the academic network described above, given a subset of papers that have been labeled as high impact papers, and/or review papers, node labeling might require, for example, predicting which among the rest of papers are also likely to be high impact papers and/or review papers. The link (label) prediction problem can be analogously defined. 

In the case of real-world multi-view networks, because each node is typically linked to only a small subset of the other nodes, a key challenge that needs to be addressed in solving the node (and link) labeling problems has to do with the sparsity of the underlying network.  A related problem has to do with the computational challenge of working with very large adjacency matrices. Network embeddings, or low-dimensional representation of each network node that summarizes the information provided about the node by the rest of the network, offers a promising approach to addressing both these problems.

\begin{definition} (Multi-view Network Embedding)
 Given a multi-view network $G$, multi-view network embedding entails learning of $d$-dimensional latent representations $X \in \Re^{|V|\times d}$, where $d << |V|$ that preserve the  structural and semantic relations among them adequately for performing one or more tasks, e.g., node label prediction.
\end{definition}
The quality of specific network embeddings (and hence that of the algorithms that produce them) have to be invariably evaluated in the context of specific applications, e.g., the predictive performance of node label predictors trained using the low-dimensional representations of nodes along with their labels, evaluated on nodes that were not part of the training data. 

The key challenge presented by multi-view network embedding over and above that of single view embedding has to do with integration of information from multiple views. Here, we can draw inspiration from multi-view learning \cite{blum1998combining,sun2013survey,xu2013survey}, where in the simplest case, each view corresponds to a different subset of features, perhaps obtained from a different modality. Multi-view learning algorithms \cite{muslea2002active+,long2008general} typically aim to maximize the agreement (with respect to the output of classifiers trained on each view, similarity of, or mutual information between low-dimensional latent representations of each view, etc). 
\section{Multi-View Network Embedding}
As noted already, our approach to solving multi-view network embedding problem leverages a single view network embedding (SVNE) method inspired by a graph soft clustering algorithm, namely, the graph factorization clustering (GFC) \cite{yu2006soft}. To solve the multi-view embedding problem, MVNE combines the information from the multiple views into the co-regularized factorization wherein the agreement between the multiple views is maximized using suitably designed objective function. MVNE combines the information from  multiple views into the co-regularized factorization space. 

\subsection{Single view network embedding}
Consider a single view network $G = (V, E)$ consisting of nodes $V$ and edges $E$. Let $K(V,U,F)$ be a bipartite graph where $U$ is a set of nodes that is disjoint from $V$ and $F$ contains all the edges connecting nodes in $V$ with nodes in $U$. Let $B = \{b_{ij}\}$ denote the $|V|\times|U|$ adjacency matrix with $b_{ij} \geq 0$ being the weight for the edge between $v_i \in V$ and $u_j \in U$. The bipartite graph $K$ induces a weight between $v_i$ and $v_j$ 
\begin{equation}
\label{eq:eq1}
w_{ij} = \sum_p b_{ip}b_{jp} =  (B \Lambda^{-1}B^T)_{ij}
\end{equation} 
where $\Lambda = diag({\lambda}_1 \ddots {\lambda}_{|U|})$ 
with ${\lambda}_p = \sum_{i} b_{ip}$ 
denotes the degree of vertex $u_p \in U$. We can normalize $W$ in Eq.(\ref{eq:eq1}) such that $\sum_{ij} w_{ij} =1$ and $w_{ij}  =p(v_i, v_j)$ according to the stationary probability of transition between $v_i$ and $v_j$ \cite{yu2006soft}. Because in a bipartite graph $K(V,U,F)$, there are no direct links between nodes in $V$, and all the paths from $v_i$ to $v_j$ must pass through nodes in $U$, we have:
\begin{equation} 
p(v_i,v_j)=p(v_i|v_j)p(v_j)
\end{equation} 
We can estimate this distribution as: $\hat{p}(v_i,v_j)=\frac{w_{ij}}{\sum_{ij}w_{ij}}$, $p(v_j)$ is given by $\frac{deg(v_j)}{\sum_{ij}w_{ij}}$ where $deg(v_j)$ represents the degree of $v_j$ and 
$p(v_i|v_j)=\sum \limits_{p=1}^{|U|}p(v_i|u_p)p(u_p|v_j)$. 
The transition probabilities between the graph $G$ and the communities $U$ (nodes of the bipartite graph) are given by $p(v_i|u_p)=\frac{b_{ip}}{\lambda_p}$ and $p(u_p|v_j)=\frac{b_{pj}}{deg(v_j)}$ where matrix $B$ denotes the weights between graph $G$ and $U$ and $\lambda_p$ denotes the degree of $u_p$.  
Hence, the transition probability between two nodes $v_i$, $v_j$ is given by:
\begin{equation} 
w_{ij} =\sum \limits_{p=1}^{d}\frac{b_{ip}b_{pj}}{\lambda_{p}}=(B\Lambda^{-1} B^T)_{ij}
\end{equation} 
Both the local and the global information in $G$ are thus encoded by matrix $B$ and diagonal matrix $\Lambda$. We can optimally preserve the information in $G$ by minimizing the objective function $\mathcal{L}(W, B\Lambda^{-1}B^T)$ where $\mathcal{L}(X,Y)=\Sigma_{ij}(x_{ij}log\frac{x_{ij}}{y_{ij}}-x_{ij}+y_{ij})$ is a variant of the K-L divergence.
Replacing $B$ by $H\Lambda$, we obtain the following objective function:
\begin{equation}
\begin{split} \label{eq:objSC}
\min_{H,\Lambda}\mathcal{L}(W,H\Lambda H^T)\\
\end{split}
\end{equation}
The objective function Eq.(\ref{eq:objSC}) is proved to be non-increasing under the update rules Eq.(\ref{eq:updateH}) and Eq.(\ref{eq:updatelambda}) for $H$ and $\Lambda$ \cite{yu2006soft}: 
\begin{equation} \label{eq:updateH}
\begin{split}
\tilde{h_{ip}} \propto &h_{ip}\Sigma_{j}log\frac{W_{ij}}{(H\Lambda H^T)_{ij}}\lambda_p h_{jp}\\
&s.t. \sum_{p=1}^d \tilde{h_{ip}}=1\\
\end{split}
\end{equation}
\begin{equation} \label{eq:updatelambda}
\begin{split}
\tilde{\lambda_p} \propto & \lambda_p\Sigma_{j}log\frac{W_{ij}}{(H\Lambda H^T)_{ij}} h_{ip} h_{jp}\\
&s.t. \sum_{p=1}^d \tilde{\lambda_{p}}=\sum_{ij} W_{ij}  
\end{split}
\end{equation}
In SVNE, the factorization $H \in \mathcal{R}^{n \times d}$ corresponds to the the single view network embedding  where $d$ is the embedding dimension. 
Because the size of the adjacency matrix representation of the network is quadratic in the number of nodes,  matrix-factorization based embedding methods typically do not scale to large networks. Hence, inspired by \cite{gemulla2011large}, we make use of more efficient encodings of the network structure: Instead of directly input the adjacent matrix,  we use a vectorized representation of adjacency matrix to perform matrix factorization.
\subsection{Multi-view Network Embedding}
Given a multi-view network $G=\{G^{(1)}, G^{(2)}, \dots G^{(k)}\}$, the key idea behind extending SVNE to  MVNE is to design the co-regularized objective function that in addition to preserving the information in each view, seeks to  maximize the agreement between the views. To accomplish this goal, we propose the following co-regularized objective function in Eq.(\ref{eq:multiobj}) which is designed to minimize the cost in each view:
\begin{small}
\begin{equation}\label{eq:multiobj}
\begin{split}
\min_{H^{(i)},\Lambda^{(i)}}& \sum_{i=1}^k  \beta_i \mathcal{L}(W^{(i)},H^{(i)}\Lambda^{(i)} {H^{(i)}}^T)\\
+&\alpha\sum_{p,q=1}^k||H^{(p)}\Lambda^{(p)}-H^{(q)}\Lambda^{(q)}||_{2}\\
&s.t. \sum_{i=1}^k  \beta_i=1
\end{split}
\end{equation}
\end{small}
Here, $H^{(i)}$ and $\Lambda^{(i)}$ represents the matrix factorization in view $i$. $\alpha$ denotes the regularization hyperparameter. $\beta_i$ is the parameter used to tune the relative importance of the different views and the role they play in maximizing the agreement between views.  If we know that some views are more informative than others, one might want to set the $\beta_i$ accordingly. In contrast, if we know that some views are likely to be noisy, we might want to deemphasize such views by setting the respective $\beta_i$ values to be small as compared to those of other views. In the absence of any information about the relative importance or reliability of the different views, we set $\beta_i$ equal to $\frac{|V^{(i)}|}{\sum_{i=1}^k |V^{(i)}|}$. 

To minimize the cost and maximize the agreement, we constrain the matrix factorization in each view to be the latent matrix factorization $H$ and $\Lambda$. This yields the objective function shown in Eq.(\ref{eq:multiobjfinal}):

\begin{equation}\label{eq:multiobjfinal}
\min_{H,\Lambda}\sum \limits_{i=1}^k \beta_i\mathcal{L}(  W^{(i)},H\Lambda H^T)\\
\end{equation}
 
We find that minimizing the objective function in Eq.(\ref{eq:multiobjfinal}) is equivalent to the following equation by ignoring the constant term:
\begin{equation}\label{eq:multiobjfinal}
\min_{H,\Lambda}\mathcal{L}( \sum \limits_{i=1}^k \beta_i W^{(i)},H\Lambda H^T)\\
\end{equation}
We co-regularize the views by choosing $\tilde{W} = \sum \limits_{i = 1} ^k \beta_iW^{(i)}$ to maximize the agreement across views. The corresponding update rules are obtained analogous to the single view case in Eq.(\ref{eq:updateH}) and  Eq.(\ref{eq:updatelambda}) by replacing $W$ with $\tilde{W}$.

\subsection*{Computational Complexity}
In the naive implementation of MVNE, each optimization iteration takes $O(d|V|^2)$ time where $|V|$ is the total number of nodes and $d$ is dimension of embedding space. However, in typical applications, $G$ is usually very sparse. In this case the time complexity of one optimization iteration using adjacency list based representation of the adjacency matrices \cite{gemulla2011large} is $O(|V|+|E|)$ (with $d$ assumed to be constant), where $|E|$ denotes the total number of edges across all of the views. 

\section{Experimental Results}
We report results of experiments designed to address the following questions:
\begin{itemize}
\item{{\bf Experiment 1:} How does SVNE compare to the state-of-the-art single view network embedding methods?}
\item{{\bf Experiment 2:} How does the MVNE algorithm introduced in this paper compare with the state-of-the-art multi-view embedding methods?} 
\item{{\bf Experiment 3:} Does MVNE embedding provide information that complements information provided by SVNE applied to the target view?}
\end{itemize}

\subsection{Experimental Setup}
\subsubsection*{Data Sets} 
\noindent{\bf Experiment 1} uses three popular single view network datasets:

\begin{itemize}
\item BlogCatalog \cite{rezasocial}: A social network of the bloggers listed on the BlogCatalog website. The labels represent blogger interests inferred through the metadata provided by the bloggers. 
\item Protein-Protein Interactions (PPI) \cite{chatr2014biogrid}: A subnetwork  of the PPI network for Homo Sapiens where the node labels correspond to biological functions of the proteins.
\item Wikipedia \cite{mahoney2011large}: This is a network of words appearing in the first million bytes of the Wikipedia dump. The labels represent the Part-of-Speech (POS) tags inferred using the Stanford POS-Tagger. 
\end{itemize}
Because each node can have multiple labels, the task entails multi-label prediction.\\

\noindent{\bf Experiments 2-3} use two multi-view network data, namely, Last.fm and Flickr \cite{bui2016labeling}:
\begin{itemize} 
  \item Last.fm: The Last.fm dataset was collected from the music network\footnote{https://www.last.fm/} with the nodes representing the users and the edges corresponding to different relationships between Last.fm users and other entities. In each view, two users are connected by an edge if they share similar interests in artists, events, etc.\cite{bui2016labeling} yielding 12 views: ArtistView (2118 nodes, 149495 links), EventView (7240 nodes, 177000 links), NeighborView (5320 nodes, 8387 links), ShoutView (7488 nodes, 14486 links), ReleaseView (4132 nodes, 129167 links), TagView (1024 nodes, 118770 links), TopAlbumView (4122 nodes, 128865 links), TopArtistView (6436 nodes, 124731 links), TopTagView (1296 nodes, 136104 links), TopTrackView (6164 nodes, 87491 links), TrackView (2680 nodes, 93358 links), and UserView (10197 nodes, 38743 links). 
  \item Flickr: The Flickr data are collected from the photo sharing website\footnote{https://www.flickr.com/}. Here, the views correspond to different aspects of Flickr (photos, comments, etc.) and edges denote shared interests between users.  For example, in the comment view, there is a link between 2 users if they have both commented on the same set of 5 or more photos. The resulting dataset has five views: CommentView (2358 nodes, 13789 links), FavoriteView (2724 nodes, 30757 links), PhotoView (4061 nodes, 91329 links), TagView (1341 nodes, 154620 links), and UserView (6163 nodes, 88052 links).
\end{itemize} 
Some basic statistics about the datasets described above are summarized in Table I.
\begin{table}
  \caption{Statical analysis of five datasets}
  \label{tab:freq}
  \begin{tabular}{llllll}
    \toprule
    Datasets&\#nodes&\#edges &\#view&\#label&\#category \\
    \midrule
    BlogCatalog & 10,312& 333,983&1&39&multi-label\\
    PPI & 3,890& 76,584&1&50&multi-label\\
    Wikipedia &4,777& 184,812&1&40&multi-label\\
    Last.fm & 10,197& 1,325,367&12&11&multi-view\\
    Flickr &6,163& 378,547&5 &10&multi-view\\
  \bottomrule
\end{tabular}
\end{table}
The results of our analyses of Last.fm and Flickr data  suggest that their node degree distributions obey the power law, a desirable property, for the application of skip-gram based models \cite{perozzi2014deepwalk}.

\subsubsection*{Parameter Tuning}

SVNE (and MVNE) are compared with other single view methods (and their multi-view extensions) using the code provided by the authors of the respective methods (with the relevant parameters set or tuned as specified in the respective papers). We explored several different settings for $d$, the dimension of the embedding space (64, 128, 256, 512) for all the methods. We used grid search over $\gamma \in \{40,80\}$ for Deepwalk and $p,q\in \{0.25, 0.50, 1, 2, 4\}$ for node2vec. 
\begin{figure*}[tp]
\centering
\includegraphics[width=0.7\textwidth]{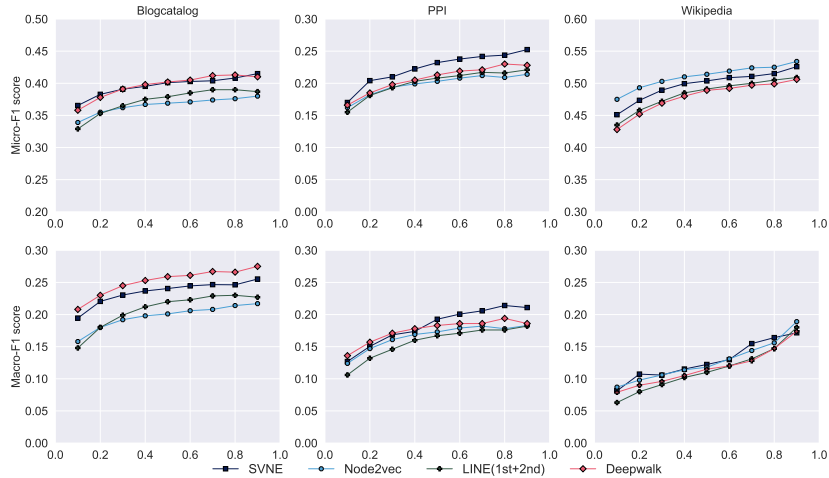}
\label{fig:temp}
\caption{{\bf SVNE compared with Deepwalk, LINE, and Node2Vec on Single View Data}. The fraction of labeled data are plotted along the x-axis. The Micro-F1 (Top) and Macro F1 (Bottom) scores are along the y-axis.}
\label{fig:singleView}
\end{figure*}

\subsubsection*{Performance Evaluation}
\noindent  In experiments 1-2, we measure the performance on the node label prediction task using different fractions of the available data (10\% to 90\% in increments of 10\%) for training and the remaining for testing the predictors. 

In experiment 3, we use 50\% of the nodes in each view for training and the rest for testing. We repeat this procedure  10 times, and report the performance (as measured by Micro F1 and Macro F1) averaged across the 10 runs. 

In each case, the embeddings are evaluated with respect to the performance of a standard one-versus-rest L2-regularized sparse logistic regression classifiers \cite{fan2008liblinear} trained to perform node label prediction.

\subsection{Exp. 1: Single view methods compared}
Experiment compares SVNE with three state-of-the-art single view embedding methods on three standard single view benchmark datasets mentioned above (Note that MVNE applied to a single view dataset yields a single view embedding):
\begin{itemize}
\item {\bf Deepwalk} which constructs a network embedding such that two nodes are close in the embedding  if the short random walks originating in the nodes are similar (i.e., generated by similar language models) \cite{perozzi2014deepwalk}.
\item {\bf LINE} which constructs a network embedding such that two nodes are close in the embedding space if their first and second order network neighborhoods are similar \cite{tang2015line}.
\item {\bf Node2Vec} which constructs a network embedding that maximizes the likelihood of preserving network neighborhoods of nodes using a biased random walk procedure to efficiently explores diverse neighborhoods \cite{grover2016node2vec}.
\end{itemize}
\subsubsection*{Results} 
The results of comparison of SVNE with Deepwalk, LINE, and Node2Vec are shown in Figure \ref{fig:singleView}. In the case of LINE, we report results for LINE(1st+2nd) (which uses 1st and 2nd order neighborhoods), in our experiments, the best performing of the 3 variants of LINE, with $d=256$. In the case of Deepwalk, we report the best results obtained with $\gamma = 40$, $w=10$, $t = 40$ and $d = 128$. For node2vec, we report the best results obtained with $p,q=1$. For SVNE, we report the results with optimal $d$, which was found to be 128 for Blogcatalog, PPI and Wikipedia. The results summarized in Figure \ref{fig:singleView} show that on Blogcatalog data, SVNE consistently outperforms Node2vec and LINE and is competitive with Deepwalk. On PPI data, SVNE outperforms all other methods in terms of Micro-F1 score and in terms of Macro-F1 when more than $50\%$ of the nodes are labeled. On wikipedia data, SVNE performs better than LINE(1st+2nd) and Deepwalk methods and is competitive with Node2vec.

\subsection{Exp. 2: MVNE compared with SVNE on Node Labeling in a Single Target View}
Experiment 2 investigates whether MVNE outperforms SVNE on node label prediction on any single target view by leveraging information from the all of the views. Considering each view of the Last.fm and Flickr data as the target view, we compare the node labeling performance using embeddings obtained using SVNE applied to the target view alone with MVNE that integrates information from all of the views. 
\begin{figure}[htp]
\begin{minipage}{1\linewidth}
  \centerline{\includegraphics[width=9cm,height=4cm]{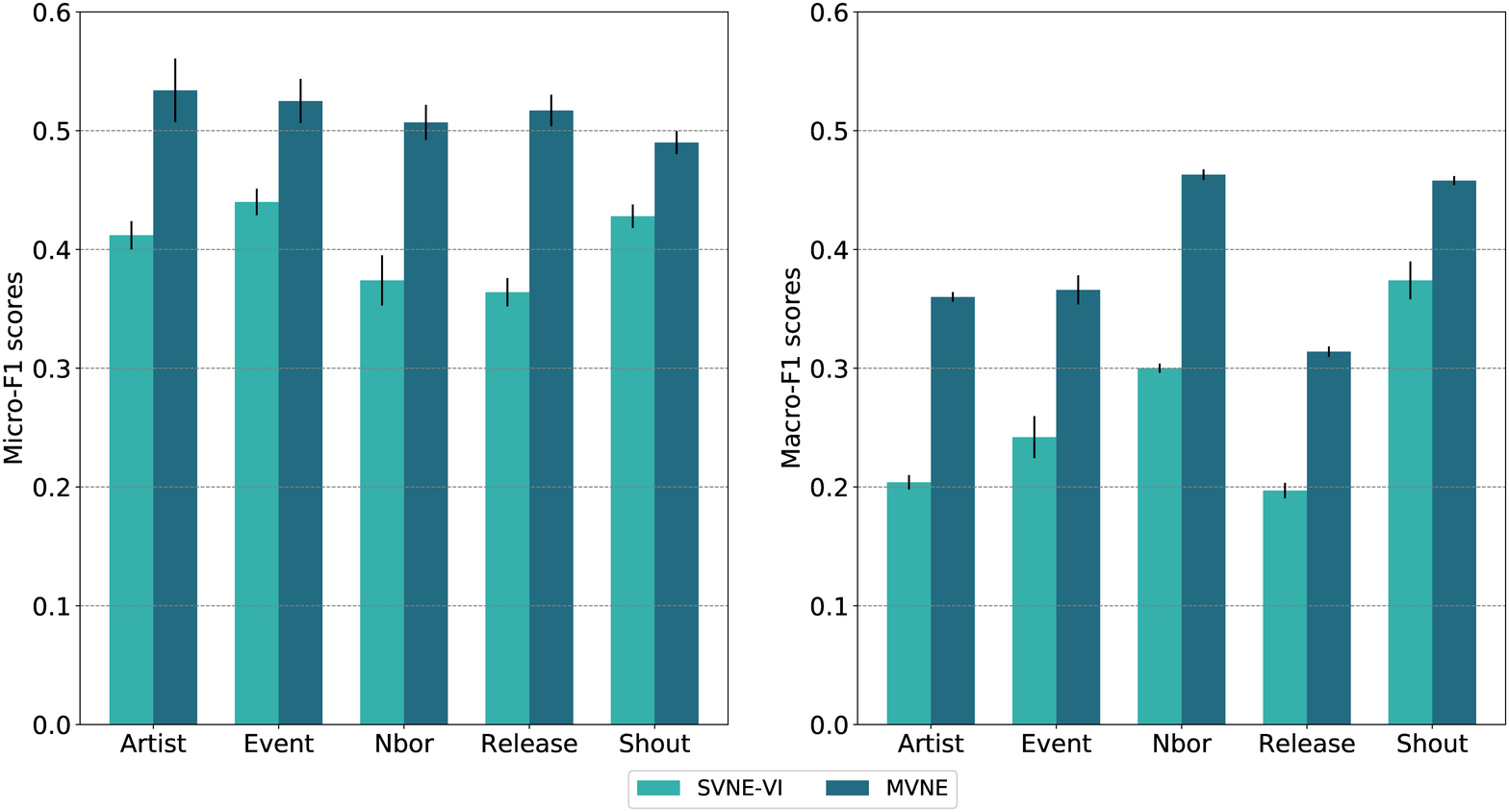}}
  \centerline{(a) Flickr}
\end{minipage}
\vfill
\begin{minipage}{1\linewidth}
  \centerline{\includegraphics[width=9cm,height=4cm]{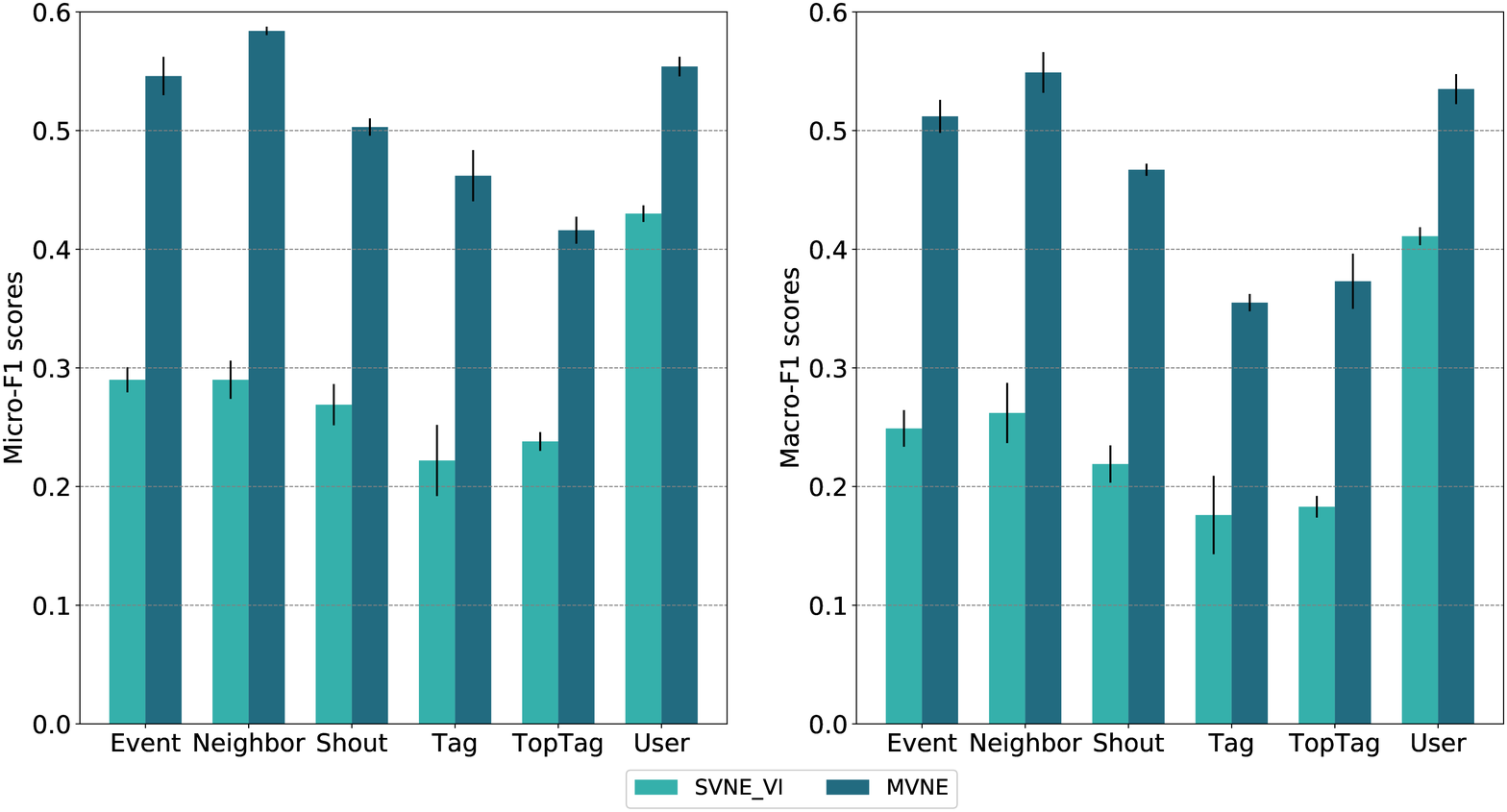}}
  \centerline{(b) Last.fm}
\end{minipage}
\caption{MVNE compared with SVNE on Flickr dataset (a) and selected six views on the Last.fm dataset (b). The view names are shown along the x-axis and Micro-F1 (Left) and Macro-F1 (Right) scores are plotted on the y-axis}
\label{fig:res}
\end{figure}
\subsubsection*{Results} 
Because of space constraints, we show only the results of comparison of MVNE with SVNE when each of the 5 views of the Flickr dataset and each of the 6 views (1 with the most nodes (Userview), one with the most edges (Event), two with most edges per node (TagView, TopTagView), and two with the  fewest edges per node(NeighborView, ShoutView)) selected from the 12 views of the Last.fm dataset are designated as the target view. The results summarized in Figure \ref{fig:res} show that MVNE consistently outperforms SVNE on each target view. We conclude that even when the goal is to predict the labels of nodes in a single target view, MVNE is able to leverage information from  all of the views to outperform SVNE applied only to the target view, by 10\% points or better. Similar results were observed with MVNE relative to SVNE when tested on the rest of the views of last.fm data (results not shown). Furthermore, similar trends were observed for all the multi-view embedding methods considered in the paper relative to their single view counterparts (results not shown). 

\section{Summary and Discussion}
We have introduced MVNE, a novel Multi-View  Network Embedding (MVNE) algorithm for constructing low-dimensional embeddings of  multi-view networks.  MVNE uses a novel objective function that maximizes the agreement between views based on both the local and global structure of the underlying multi-view network.  We have shown that (i) SVNE, the single view version of MVNE, is competitive with or superior to the state-of-the-art single view network embedding methods when the embeddings are used for labeling unlabeled nodes in the networks; (ii) MVNE substantially outperforms single view methods on integrated view, as well as state-of-the-art multi-view graph methods for aggregating information from multiple views, when the embeddings are used for labeling nodes in multi-view networks; and (iii)  MVNE outperforms SVNE, when used to predict node labels in any target view, suggesting that it is able to effectively integrate from all of the views, information that is useful for labeling nodes in the target view. 

\subsection{Related work}
There is a growing body of recent works on multi-view learning algorithms, e.g., \cite{liu2013multi,wu2014multi,ma2017multi}, that attempt to integrate information across the multiple views to optimize the predictive performance of the classifier (see \cite{xu2013survey,sun2013survey}). Some multi-view learning methods seek to maximize the agreement between views using regularization \cite{sindhwani2005co,kumar2011core} whereas others seek to optimally selecting subsets of features from different views for each prediction task \cite{ liu2013multi,lu2018learning}  However, these methods were not designed for network embedding. Most of the existing multi-view learning algorithms are either not directly applicable to multi-view networks or are not designed to cope with high degrees of data sparsity, a key challenge in modeling real-world multi-view networks.

Network embedding methods aim to produce information preserving low-dimensional embeddings of nodes in large networks. State-of-the-art network embedding methods include Deepwalk \cite{perozzi2014deepwalk}, LINE \cite{tang2015line} and node2vec \cite{grover2016node2vec} are limited to single view networks, i.e, networks with a single type of links. However, most real-world networks are comprised of multiple types of nodes and links \cite{dickison2016multilayer,tang2015line,berlingerio2013multidimensional} wherein each type of link induces a view. Hence, there is a growing interest in network embedding methods for multi-view networks \cite{kivela2014multilayer,bazzi2016community,scott2011social,bui2016labeling}. Some multi-view network embedding methods use canonical correlation analysis (CCA)\cite{andrew2013deep,wang2015deep,benton2016learning} to integrate information from multiple views. Others construct multi-view embeddings by  integrating   embeddings obtained from the individual views. Examples include  MVWE \cite{qu2017attention} which uses a weighted voting mechanism to combine information from multiple views; MVE2vec \cite{shi2018mvn2vec} which attempts to balance the preservation of unique information provided by specific views against information that is shared by multiple views; and DMNE \cite{ni2018co} which uses a co-regularized cost function to combine information from different views. MVWE, MVE2vec, and DMNE use deep neural network models at their core. Specifically, MVWE and MVE2vec are based on a skip-gram model and DMNE is based on an AutoEncoder. 

In contrast to the existing multi-view network embedding methods, MVNE exploits a recently discovered connection between network adjacency matrix factorization and network embedding \cite{qiu2018network} to utilize GFC \cite{yu2006soft}, a graph factorization method, to perform single view network embedding. MVNE extends the resulting single view network embedding algorithm to the multi-view setting. Inspired by \cite{lai2017prune},  MVNE uses a novel objective function that maximizes the agreement between views while combining information derived from the local as well as the global structure of the underlying multi-view networks.  Like DMNE \cite{ni2018co},  MVNE uses a co-regularized objective function to maximize the agreement in the embedding space and to control the embedding dimension. Unlike DMNE which requires on computationally expensive training of a deep neural network, MVNE is considerably more efficient and hence scalable to large networks. 

\subsection{Future Directions}
Work in progress is aimed at extending MVNE (i) to cope with dynamic update of graphs e.g., using asynchronous stochastic gradient descent (SGD) to update the latent space with the only newly added or deleted edges or nodes;  and (ii) work with multi-modal networks that include richly structured digital objects (text, images, videos, etc).

\section*{Acknowledgements}
This project was supported in part by the National Center for Advancing Translational Sciences, National Institutes of Health through the grant UL1 TR000127 and TR002014, by the National Science Foundation, through the grants 1518732, 1640834, and 1636795; the Pennsylvania State University’s Institute for Cyberscience and the Center for Big Data Analytics and Discovery Informatics; the Edward Frymoyer Endowed Professorship in Information Sciences and Technology at Pennsylvania State University and the Sudha Murty Distinguished Visiting Chair in Neurocomputing and Data Science funded by the Pratiksha Trust at the Indian Institute of Science [both held by Vasant Honavar]. The content is solely the responsibility of the authors and does not necessarily represent the official views of the sponsors.


\begin{thebibliography}{10}

\bibitem{andrew2013deep}
G.~Andrew, R.~Arora, J.~Bilmes, and K.~Livescu.
\newblock Deep canonical correlation analysis.
\newblock In {\em International Conference on Machine Learning}, pages
  1247--1255, 2013.

\bibitem{bazzi2016community}
M.~Bazzi, M.~A. Porter, S.~Williams, M.~McDonald, D.~J. Fenn, and S.~D.
  Howison.
\newblock Community detection in temporal multilayer networks, with an
  application to correlation networks.
\newblock {\em Multiscale Modeling \& Simulation}, 14(1):1--41, 2016.

\bibitem{benton2016learning}
A.~Benton, R.~Arora, and M.~Dredze.
\newblock Learning multiview embeddings of twitter users.
\newblock In {\em Proceedings of the 54th Annual Meeting of the Association for
  Computational Linguistics}, volume~2, pages 14--19, 2016.

\bibitem{berlingerio2013multidimensional}
M.~Berlingerio, M.~Coscia, F.~Giannotti, A.~Monreale, and D.~Pedreschi.
\newblock Multidimensional networks: foundations of structural analysis.
\newblock {\em World Wide Web}, 16(5-6):567--593, 2013.

\bibitem{blum1998combining}
A.~Blum and T.~Mitchell.
\newblock Combining labeled and unlabeled data with co-training.
\newblock In {\em Proceedings of the eleventh annual conference on
  Computational learning theory}, pages 92--100. ACM, 1998.

\bibitem{bui2016labeling}
N.~Bui, T.~Le, and V.~Honavar.
\newblock Labeling actors in multi-view social networks by integrating
  information from within and across multiple views.
\newblock In {\em Big Data (Big Data), 2016 IEEE International Conference on},
  pages 616--625. IEEE, 2016.

\bibitem{cao2014tensor}
B.~Cao, L.~He, X.~Kong, S.~Y. Philip, Z.~Hao, and A.~B. Ragin.
\newblock Tensor-based multi-view feature selection with applications to brain
  diseases.
\newblock In {\em Data Mining (ICDM), 2014 IEEE International Conference on},
  pages 40--49. IEEE, 2014.

\bibitem{chatr2014biogrid}
A.~Chatr-Aryamontri, B.-J. Breitkreutz, R.~Oughtred, L.~Boucher, S.~Heinicke,
  D.~Chen, C.~Stark, A.~Breitkreutz, N.~Kolas, L.~O'donnell, et~al.
\newblock The biogrid interaction database: 2015 update.
\newblock {\em Nucleic acids research}, 43(D1):D470--D478, 2014.

\bibitem{chaudhuri2009multi}
K.~Chaudhuri, S.~M. Kakade, K.~Livescu, and K.~Sridharan.
\newblock Multi-view clustering via canonical correlation analysis.
\newblock In {\em Proceedings of the 26th annual international conference on
  machine learning}, pages 129--136. ACM, 2009.

\bibitem{cui2017survey}
P.~Cui, X.~Wang, J.~Pei, and W.~Zhu.
\newblock A survey on network embedding.
\newblock {\em arXiv preprint arXiv:1711.08752}, 2017.

\bibitem{dhillon2011multi}
P.~Dhillon, D.~P. Foster, and L.~H. Ungar.
\newblock Multi-view learning of word embeddings via cca.
\newblock In {\em Advances in Neural Information Processing Systems}, pages
  199--207, 2011.

\bibitem{dickison2016multilayer}
M.~E. Dickison, M.~Magnani, and L.~Rossi.
\newblock {\em Multilayer social networks}.
\newblock Cambridge University Press, 2016.

\bibitem{elkahky2015multi}
A.~M. Elkahky, Y.~Song, and X.~He.
\newblock A multi-view deep learning approach for cross domain user modeling in
  recommendation systems.
\newblock In {\em Proceedings of the 24th International Conference on World
  Wide Web}, pages 278--288. International World Wide Web Conferences Steering
  Committee, 2015.

\bibitem{fan2008liblinear}
R.-E. Fan, K.-W. Chang, C.-J. Hsieh, X.-R. Wang, and C.-J. Lin.
\newblock Liblinear: A library for large linear classification.
\newblock {\em Journal of machine learning research}, 9(Aug):1871--1874, 2008.

\bibitem{gemulla2011large}
R.~Gemulla, E.~Nijkamp, P.~J. Haas, and Y.~Sismanis.
\newblock Large-scale matrix factorization with distributed stochastic gradient
  descent.
\newblock In {\em Proceedings of the 17th ACM SIGKDD international conference
  on Knowledge discovery and data mining}, pages 69--77. ACM, 2011.

\bibitem{grover2016node2vec}
A.~Grover and J.~Leskovec.
\newblock node2vec: Scalable feature learning for networks.
\newblock In {\em Proceedings of the 22nd ACM SIGKDD international conference
  on Knowledge discovery and data mining}, pages 855--864. ACM, 2016.

\bibitem{kivela2014multilayer}
M.~Kivel{\"a}, A.~Arenas, M.~Barthelemy, J.~P. Gleeson, Y.~Moreno, and M.~A.
  Porter.
\newblock Multilayer networks.
\newblock {\em Journal of complex networks}, 2(3):203--271, 2014.

\bibitem{kumar2011core}
A.~Kumar, P.~Rai, and H.~Daume.
\newblock Co-regularized multi-view spectral clustering.
\newblock In {\em Advances in neural information processing systems}, pages
  1413--1421, 2011.

\bibitem{lai2017prune}
Y.-A. Lai, C.-C. Hsu, W.~H. Chen, M.-Y. Yeh, and S.-D. Lin.
\newblock Prune: Preserving proximity and global ranking for network embedding.
\newblock In {\em Advances in Neural Information Processing Systems}, pages
  5263--5272, 2017.

\bibitem{li2010two}
G.~Li, S.~C. Hoi, and K.~Chang.
\newblock Two-view transductive support vector machines.
\newblock In {\em Proceedings of the 2010 SIAM International Conference on Data
  Mining}, pages 235--244. SIAM, 2010.

\bibitem{liu2013multi}
J.~Liu, C.~Wang, J.~Gao, and J.~Han.
\newblock Multi-view clustering via joint nonnegative matrix factorization.
\newblock In {\em Proceedings of the 2013 SIAM International Conference on Data
  Mining}, pages 252--260. SIAM, 2013.

\bibitem{long2008general}
B.~Long, P.~S. Yu, and Z.~Zhang.
\newblock A general model for multiple view unsupervised learning.
\newblock In {\em Proceedings of the 2008 SIAM international conference on data
  mining}, pages 822--833. SIAM, 2008.

\bibitem{lu2018learning}
C.-T. Lu, L.~He, H.~Ding, B.~Cao, and S.~Y. Philip.
\newblock Learning from multi-view multi-way data via structural factorization
  machines.
\newblock In {\em Proceedings of the 2018 World Wide Web Conference on World
  Wide Web. International World Wide Web Conferences Steering Committee}, pages
  1593--1602, 2018.

\bibitem{lu2017multilinear}
C.-T. Lu, L.~He, W.~Shao, B.~Cao, and P.~S. Yu.
\newblock Multilinear factorization machines for multi-task multi-view
  learning.
\newblock In {\em Proceedings of the Tenth ACM International Conference on Web
  Search and Data Mining}, pages 701--709. ACM, 2017.

\bibitem{ma2017multi}
G.~Ma, L.~He, C.-T. Lu, W.~Shao, P.~S. Yu, A.~D. Leow, and A.~B. Ragin.
\newblock Multi-view clustering with graph embedding for connectome analysis.
\newblock In {\em Proceedings of the 2017 ACM on Conference on Information and
  Knowledge Management}, pages 127--136. ACM, 2017.

\bibitem{mahoney2011large}
M.~Mahoney.
\newblock Large text compression benchmark, 2011.

\bibitem{muslea2002active+}
I.~Muslea, S.~Minton, and C.~A. Knoblock.
\newblock Active+ semi-supervised learning= robust multi-view learning.
\newblock In {\em ICML}, volume~2, pages 435--442, 2002.

\bibitem{ni2018co}
J.~Ni, S.~Chang, X.~Liu, W.~Cheng, H.~Chen, D.~Xu, and X.~Zhang.
\newblock Co-regularized deep multi-network embedding.
\newblock In {\em Proceedings of the 2018 World Wide Web Conference on World
  Wide Web}, pages 469--478. International World Wide Web Conferences Steering
  Committee, 2018.

\bibitem{perozzi2014deepwalk}
B.~Perozzi, R.~Al-Rfou, and S.~Skiena.
\newblock Deepwalk: Online learning of social representations.
\newblock In {\em Proceedings of the 20th ACM SIGKDD international conference
  on Knowledge discovery and data mining}, pages 701--710. ACM, 2014.

\bibitem{qiu2018network}
J.~Qiu, Y.~Dong, H.~Ma, J.~Li, K.~Wang, and J.~Tang.
\newblock Network embedding as matrix factorization: Unifying deepwalk, line,
  pte, and node2vec.
\newblock In {\em Proceedings of the Eleventh ACM International Conference on
  Web Search and Data Mining}, pages 459--467. ACM, 2018.

\bibitem{qu2017attention}
M.~Qu, J.~Tang, J.~Shang, X.~Ren, M.~Zhang, and J.~Han.
\newblock An attention-based collaboration framework for multi-view network
  representation learning.
\newblock In {\em Proceedings of the 2017 ACM on Conference on Information and
  Knowledge Management}, pages 1767--1776. ACM, 2017.

\bibitem{rezasocial}
Z.~Reza and L.~Huan.
\newblock Social computing data repository.

\bibitem{scott2011social}
J.~Scott.
\newblock Social network analysis: developments, advances, and prospects.
\newblock {\em Social network analysis and mining}, 1(1):21--26, 2011.

\bibitem{shi2018mvn2vec}
Y.~Shi, F.~Han, X.~He, C.~Yang, J.~Luo, and J.~Han.
\newblock mvn2vec: Preservation and collaboration in multi-view network
  embedding.
\newblock {\em arXiv preprint arXiv:1801.06597}, 2018.

\bibitem{sindhwani2005co}
V.~Sindhwani, P.~Niyogi, and M.~Belkin.
\newblock A co-regularization approach to semi-supervised learning with
  multiple views.
\newblock In {\em Proceedings of ICML workshop on learning with multiple
  views}, pages 74--79, 2005.

\bibitem{sun2013survey}
S.~Sun.
\newblock A survey of multi-view machine learning.
\newblock {\em Neural Computing and Applications}, 23(7-8):2031--2038, 2013.

\bibitem{tang2015line}
J.~Tang, M.~Qu, M.~Wang, M.~Zhang, J.~Yan, and Q.~Mei.
\newblock Line: Large-scale information network embedding.
\newblock In {\em Proceedings of the 24th International Conference on World
  Wide Web}, pages 1067--1077. International World Wide Web Conferences
  Steering Committee, 2015.

\bibitem{wang2015deep}
W.~Wang, R.~Arora, K.~Livescu, and J.~Bilmes.
\newblock On deep multi-view representation learning.
\newblock In {\em Proceedings of the 32nd International Conference on Machine
  Learning (ICML-15)}, pages 1083--1092, 2015.

\bibitem{wu2014multi}
J.~Wu, Z.~Hong, S.~Pan, X.~Zhu, Z.~Cai, and C.~Zhang.
\newblock Multi-graph-view learning for graph classification.
\newblock In {\em Data Mining (ICDM), 2014 IEEE International Conference on},
  pages 590--599. IEEE, 2014.

\bibitem{xu2013survey}
C.~Xu, D.~Tao, and C.~Xu.
\newblock A survey on multi-view learning.
\newblock {\em arXiv preprint arXiv:1304.5634}, 2013.

\bibitem{yu2006soft}
K.~Yu, S.~Yu, and V.~Tresp.
\newblock Soft clustering on graphs.
\newblock In {\em Advances in neural information processing systems}, pages
  1553--1560, 2006.

\end{thebibliography}
\end{document}